\documentclass{article}
\usepackage{spconf,amsmath,graphicx}
\usepackage{float}
\usepackage{subfigure}
\usepackage{hyperref}
\hypersetup{hypertex=true,
colorlinks=true,
linkcolor=blue,
anchorcolor=blue,
citecolor=blue}

\title{Towards Radar Emitter Recognition in Changing Environments \\with Domain Generalization}
\name{Honglin Wu, Xueqiong Li, Long Lan, Liyang Xu, Yuhua Tang\thanks{Thanks to National Natural Science Foundation of China (No. 62101575 and No.12002380), the National University of Defense Technology Foundation (No.ZK22-57 and No. ZK20-52) for funding.}}
\address{Institute for Quantum Information \& State Key Laboratory of High Performance Computing, \\College of Computer Science and Technology, National University of Defense Technology, \\Changsha, China}
\begin{document}
%
\maketitle
\begin{abstract}
Analyzing radar signals from complex Electronic Warfare (EW) environment is a non-trivial task. However, in the real world, the changing EW environment results in inconsistent signal distribution, such as the pulse repetition interval (PRI) mismatch between different detected scenes. In this paper, we propose a novel domain generalization framework to improve the adaptability of signal recognition in changing environments. Specifically, we first design several noise generators to simulate varied scenes. Different from conventional augmentation methods, our introduced generators carefully enhance the diversity of the detected signals and meanwhile maintain the semantic features of the signals. Moreover, we propose a signal scene domain classifier that works in the manner of adversarial learning. The proposed classifier guarantees the signal predictor to generalize to different scenes. Extensive comparative experiments prove the proposed method’s superiority.

\end{abstract}

\section{Introduction}
\label{sec:intro}
\vspace{-0.2cm}

In the existing literature \cite{2006ELINT, 1990Radar, 2018LiCNN, 2020LiRNN}, radar signals are described using Pulse Description Words (PDWs), which contain several statistical features, such as pulse width (PW), carrier frequency (CF), pulse amplitude (PA), time of arrival (TOA), and direction of arrival (DOA)~\cite{2006ELINT}. Among these statistical features, TOA is an easily captured one whose first-order difference is pulse repetition interval (PRI). PRI is a principal parameter for radar emitter recognition and PRI sequences often indicate the working pattern of radar emitters.

\begin{figure}[t]

\begin{minipage}[b]{1.0\linewidth}
  \centering
  \centerline{\includegraphics[width=8cm]{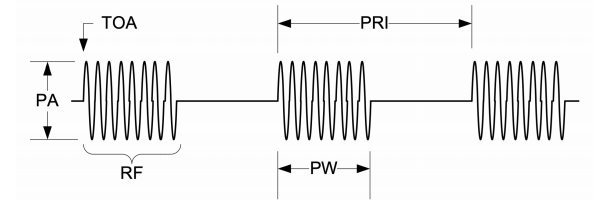}}
  \vspace{-0.3cm}
  \centerline{(a) Radar signal pulse description words}\medskip
\end{minipage}

\begin{minipage}[b]{1.0\linewidth}
  \centering
  \centerline{\includegraphics[width=8.5cm]{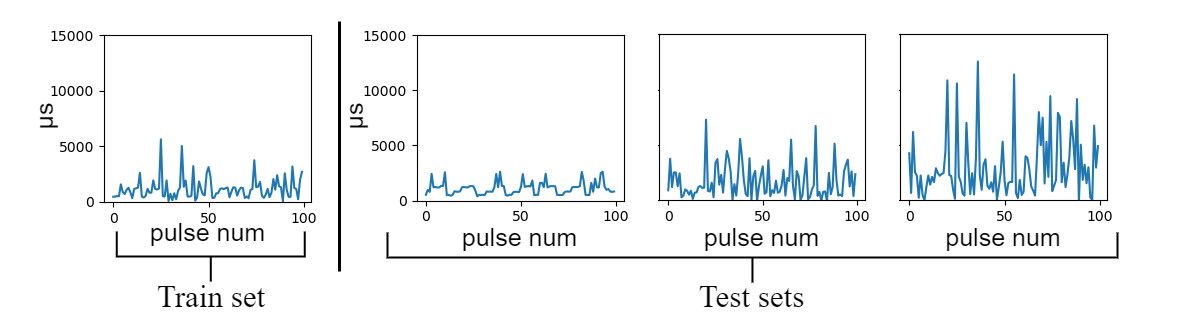}}
  \centerline{(b) Radar PRI sequence in real EW environment}\medskip
\end{minipage}
\vspace{-0.9cm}
\caption{(a) pulse description words, with PRI being the principal parameter. (b) shows the PRI sequence with high ratios of missing and spurious pulse from training and testing sets in different EW environments.}
\vspace{-0.5cm}
\label{fig:res}
\end{figure}
 
With the rapid development and usage of radar systems, the EW environment is becoming increasingly complex, posing severe challenges to recognizing radar emitters \cite{icasspRadar2}. The high ratios of missing and spurious pulses in the pulse streams destroy the PRI regularity to a great extent. Besides, measurement errors may occur when receiving the signals, which along with noise, significantly increase the emitter signal recognition complexity.

Deep learning models have been widely used for radar emitter recognition, allowing complex signal data recognition \cite{2018LiCNN, 2020LiRNN, icasspRadar}. A fundamental assumption underlying the remarkable success is that the test data follows similar statistics as the training data. However, in a real EW environment, error ratios of the signal to be recognized are often fixed at a value different from the detected signal. This is due to man-made interference and environmental changes, which causes mismatches between the training and testing sets. Therefore, a significant yet seldom investigated problem arises: Can a radar signal be accurately recognized when the electromagnetic environment changes and produce a different error rate? 

We found that if a model is trained on historical signal data and is used to identify newly received signal data, the performance will seriously deteriorate. Furthermore, in practice, collecting signals for all cases is impossible so the signal data cannot be divided into multiple training domains. We investigate how to recognize a radar emitter in future-changing environments using a model trained on captured data. Therefore, the task is considered a single-domain generalization (DG) problem.

This paper introduces a novel framework to solve this challenging task. To summarize, this work's main contributions are as follows: 
\vspace{-0.2cm}
\begin{itemize}
\item We propose a novel domain generalization network for radar emitter signal recognition. Following the pattern of adversarial learning, the proposed method can significantly improve the recognition performance of radar emitters under changing EW environments.
\vspace{-0.2cm}

\item In our method, we proposed to use several generators to simulate the noise of changing environments, which ensures the generalization of different scenes for the challenging recognition task.
\vspace{-0.2cm}

\item Extensive experiments on radar emitter recognition demonstrate our method's superior performance. Specifically, the proposed method attains up to 4.9\% and 9.45\% improvement compared with existing DG and emitter recognition methods, respectively.
\vspace{-0.2cm}
\end{itemize}

\vspace{-0.3cm}
\section{Related Works}
\label{sec:Related Works}
\vspace{-0.2cm}

{\bf Radar Emitter Signal Recognition:}
Conventional feature-based methods recognize radar emitter signals based on the PRI's histogram \cite{1990Radar}, posing an inefficient and untrustworthy solution. In recent years, deep learning has become a key technology for radar signal recognition. For instance, \cite{2018LiCNN} used several neural networks, including MLP, CNN, and LSTM, to recognize radar signals. In \cite{2020LiRNN}, the authors proposed an attention-based RNN to recognize radar signals, which was effective on time-series data. The work of \cite{2020ACSE} proposed an Asymmetric Convolution Squeeze-and-Excitation (ACSE) network, which achieved a high recognition accuracy under a low signal-to-noise ratio using the normalized autocorrelation features as input. In \cite{2021TCN}, the authors used a temporal convolution network (TCN) for radar signal recognition, which provided robust performance even under missing and spurious pulses. However, current works did not consider the inconsistent signal distribution caused by environmental variations and electronic jamming.

\noindent {\bf Domain Generalization:}
Domain Generalization (DG) aims to learn the model with data from the source domain, allowing it to generalize unseen domains \cite{2021Generalizing}. Data manipulation and representation learning are two major methods, in which the data manipulation method involves two popular techniques, i.e., data augmentation \cite{2017Peng,2019Khirodkar,2018Tremblay,2018Shankar,2018Volpi} and data generation \cite{2014VAE,2016GAN,2017mixup,icasspDG}. 
Both methods focus on manipulating the inputs to assist in learning general representations and have achieved promising performance on popular benchmarks while remaining conceptually and computationally simple. Representation learning involves two representative techniques: domain-invariant representation learning \cite{2019d-SNE,2020MADA,2018Li,2019IRM} and feature disentanglement \cite{2020Piratla,2019DIVA}. Those methods learn domain invariant representations or disentangle the features into domain-shared or domain-specific parts for better generalization, which are the most popular in domain generalization. However, due to various particularities of radar signal data and its features, not every DG method is suitable for radar emitter signal recognition.

\begin{figure*}
\begin{minipage}[b]{1.0\linewidth}
  \centering
  \centerline{\includegraphics[width=17cm]{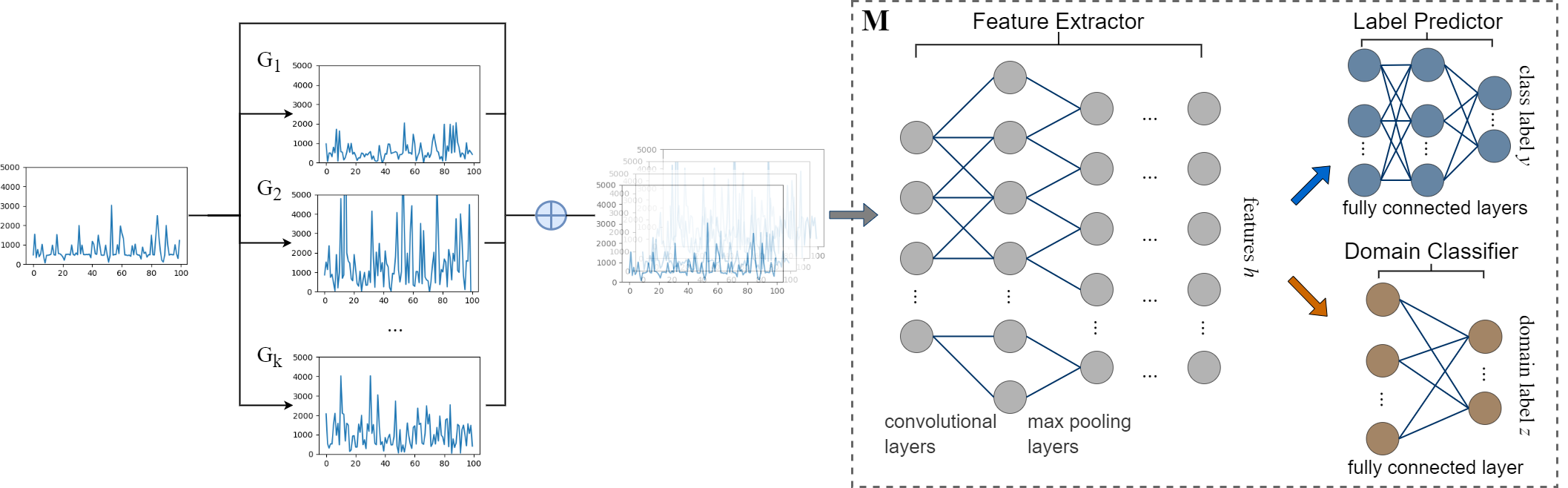}}
\end{minipage}
\vspace{-0.75cm}
\caption{Overall architecture of our method. {\it G} is the noise generator that simulates signals in different scenes. Different {\it G} share the same structure but have different weights. The recognition model {\it M} includes a feature extractor, a label predictor, and a domain classifier.}
\vspace{-0.3cm}
\label{fig:model}
\end{figure*}

\vspace{-0.3cm}
\section{Method}
\label{sec:Method}
\vspace{-0.2cm}

The architecture of the proposed radar emitter recognition method is illustrated in  Fig.\ref{fig:model}. Overall, the framework includes noise generators {\it G} and a signal recognition model {\it M}. Specifically, noise generators {\it G} are first utilized to simulate signals in different scenes. Different {\it G} share the same structure but have different weights. The recognition model {\it M} works in the manner of adversarial learning. {\it M} learns the cross-domain feature of signals from different environments.

The target of the method is to train with source domain signals $\mathcal{S}$ and then generalize it to the unseen target domain $\mathcal{T}$. Suppose the source domain $\mathcal{S} = \{x_i, y_i\}^{N_s}_{i=1}$, the target domain $\mathcal{T} = \{x_i, y_i\}^{N_t}_{i=1}$, where $x_i$, $y_i$ is the $i^{th}$ emitter PRI sequence and class label, respectively. $N_s$, $N_t$ represent the number of samples in the source and target domain, respectively.

\vspace{-0.3cm}
\subsection{Noise Generator {\it G}}
\vspace{-0.2cm}

G can convert the detected radar PRI sequence $x$ to a new PRI sequence $x^+$:
\vspace{-0.2cm}
\begin{gather}
    x^+ = G(x),\\
    \mathcal{S}^+ = \{(G(x_i),y_i)|(x_i, y_i)\in \mathcal{S}\},
\end{gather}
where $x^+$ has the same emitter feature as $x$, but the environmental error of $x^+$ and $x$ is different, as the latter mainly contains missing and spurious pulses with a small amount of measurement error.

The noise generator {\it G} is only used in training as preprocessing. Given that we aim to simulate signals in various environments through the known signals, {\it G}  generates new noise signal PRI sequences through special augmentation operations. 
In this work, we employ three sub-operations: add pulses, drop pulses, and Gaussian noise. During training, an operation including random sub-operations is performed on all data in training set $\mathcal{S}$ to generate a new set $\mathcal{{S}^+}$.

\vspace{-0.3cm}
\subsection{Signal Recognition Model {\it M}}
\vspace{-0.2cm}

There are three modules in {\it M}: (i) Feature Extractor {\it F}:$\mathcal X$ $\rightarrow$ $\mathcal H$, where $\mathcal X$ is the signal space and $\mathcal H$ is the feature space. {\it F} includes four convolution layers followed by max pooling and activation layers, which output a feature vector of the input signal data. (ii) Label Classifier {\it C}: $\mathcal H$ $\rightarrow$ $\mathcal Y$, where $\mathcal Y$ is the label space. {\it C} includes three fully connected layers followed by activation layers, and its task is to classify the emitter. (iii) Domain classifier {\it D}: $\mathcal H$ $\rightarrow$ $\mathcal Z$, where $\mathcal Z$ is domain space. {\it D} includes one fully connected layer followed by softmax. {\it D} is used to distinguish whether the signal data originate from the same error environment.

\subsection{Model Optimization}
\vspace{-0.2cm}

For the recognition model {\it M}, given a minibatch $\mathcal{B} = \{\mathbf{x}, \mathbf{y}\} \in \mathcal{S}\bigcup \mathcal{S}^+$, where $x$ is source PRI sequence, $x^+$ is the synthetic PRI sequences originating from $x$, and $\mathbf{y}$ is the class label, and the domain label $\mathbf{z}$, {\it M} is optimized by:

\vspace{-0.6cm}
\begin{gather}
    \mathcal{L}_{ce}(\mathbf{y_i}, \mathbf{\hat{y_i}}) = - \sum_m y_i^m \log(\hat{y_i^m}),
    \\
    \mathcal{L}_M = \mathcal{L}_{ce}(\mathbf{y_i}, \mathbf{\hat{y_i}}) 
    + \alpha \| \mathbf{h}-\mathbf{h^+} \| _2^2
    - \beta \mathcal{L}_{ce}(\mathbf{z_i}, \mathbf{\hat{z_i}}),
    \label{eq}
\end{gather}
where $\mathbf{\hat{y_i}} = C(F(\mathbf{x_i}))$, 
$\mathbf{\hat{z_i}} = D(F(\mathbf{x_i}))$, 
$\mathbf{h}= F(\mathbf{x})$, $\mathbf{h^+}= F(\mathbf{x^+})$, 
$y_i^m$ and $\hat{y_i^m}$ respectively represent the $m^{th}$ dimension of $\mathbf{y_i}$ and $\mathbf{\hat{y_i}}$, $\mathcal{L}_{ce}$ is the cross-entropy loss used for label and domain classification. 
The second term forces {\it F} to learn the domain invariant representation between $\mathcal{S}$ and $\mathcal{S}^+$ in the embedding space, and $\alpha$ and $\beta$ are two hyper-parameter to balance the loss. In our task, we force the model to classify the emitters better, but it is challenging to distinguish which environment the emitters are from. Through adversarial learning, by minimizing Eq \ref{eq}, we improve the model's generalization ability for radar emitter recognition.

\vspace{-0.3cm}
\subsection{Data Augmentation in Radar Signal Recognition}
\vspace{-0.2cm}

To generate representative radar signals for different domains, we designed different noise generators accordingly as indicated in Fig.~\ref{fig:model}. There are many useful data augmentation methods in the image processing area, such as crop, translation, flip, and rotation. Those methods are proved to be very efficient in constructing training examples as they improve the diversities of the data. However, the above traditional data augmentation methods do not work as well for the radar signal recognition task since they could induce noises to the signal, leading to unexpected recognition results. For example, shifting the signal curve means changing the corresponding PRI parameter, which stands for different radar signals. In our model, we carefully designed radar signal data augmentation methods that simulate the noise EW environment and keep the semantic feature of radar signals. As we consider several different noise generators in the radar signal data augmentation, they significantly enrich the training examples in different noise environments and thus effectively relieve the mismatch between training and testing sets.

\vspace{-0.2cm}
\section{Experiments}
\label{sec:Experiments}

\vspace{-0.25cm}
\subsection{Experimental Setup}
\vspace{-0.2cm}

As mentioned above, PRI is the principal parameter for radar emitter recognition. There are six basic types of PRI modulation: constant PRI, jittered PRI, sliding PRI, wobulated PRI, staggered PRI, dwell and switch PRI. Each PRI category has different formulations associated with different radar functions. In most cases, the remaining parameters in PDWs are used to assist PRI for emitter recognition. Therefore, this paper employs PRI sequences for radar emitter recognition. 

{\noindent\bf Dataset and Evaluation:}
There is rarely a public dataset available for radar emitter recognition due to the confidentiality of radar emitters. Therefore, we create a standard dataset according to the task requirement and the typical PRI range. The dataset contains PRI sequences of 10 emitters, six of which have different types of PRI modulation, and the remaining four are staggered PRI but with different PRI values. 
Besides, in our experiments, each sequence is assumed to drop a certain probability by $\rho_m$. $\rho_m = \frac{\sum_i a_i}{\sum_i a_i + b_i}$, where $a_i$ is the number of lost pulses in the $i^{th}$ period, and $b_i$ are the remaining pulses in the $i^{th}$ period. Spurious pulses are added between two adjacent pulses with their number subjecting to a Poisson distribution with a mean of $\rho_n(1 - \rho_m)$. In this way, the noise number to pulse number ratio in the streams is guaranteed to be, on average, $\rho_n$. Additionally, measurement errors are added to the sequences following a Gaussian distribution of $\rho_r$ ratio. In the training set, $P_{train}: (\rho_r, \rho_m, \rho_n)$ is set as $P_{train}:(0.05, 0.2, 0.4)$. In the testing set, the signal data is divided into four groups, where the error ratios are $P_1:(0.02, 0.05, 0.2)$ , $P_2:(0.05, 0.2, 0.4)$ , $P_3:(0.05, 0.3, 0.6)$ and $P_4:(0.1, 0.5, 0.8)$. The training set considers the emitter signal in typical EW environments, and the testing sets are of various EW environments that cover the best to the worst cases in real scenarios.
We calculate the model's recognition accuracy for emitters 1) in different EW environments; 2) of different PRI modulation; 3) of the same PRI modulation but with different PRI values.

\vspace{-0.3cm}
\subsection{Experimental Results}
\vspace{-0.2cm}

\begin{table}[t]
\centering
\vspace{-0.25cm}
\caption{Experimental results on the created testing set. The best results of each EW environment are bold.}
\setlength{\tabcolsep}{3mm}{
\begin{tabular}{llllll}
\hline
Methods                     & $P_1$     & $P_2$     & $P_3$     & $P_4$     & Avg.      \\
\hline
CNN \cite{2018LiCNN}        & 85.6      & 92.5      & 73.2      & 48.2      & 74.9      \\
A-RNN \cite{2020LiRNN}      & 86.2      & 93.7      & 75.1      & 50.5      & 76.4      \\
ACSE \cite{2020ACSE}        & 86.6      & 93.1      & 74.9      & 49.0      & 75.9      \\
TCN  \cite{2021TCN}         & 87.0      & {\bf94.2} & 74.6      & 52.2      & 77.0      \\
ERM \cite{ERM}              & 88.2      & 93.3      & 79.3      & 57.4      & 79.6      \\
d-SNE \cite{2019d-SNE}      & 89.5      & 93.6      & 81.6      & 61.4      & 81.6      \\
MADA  \cite{2020MADA}       & 88.9      & 93.1      & 78.6      & 59.7      & 80.1      \\
Ours                        & {\bf92.7} & 94.0      & {\bf86.5} & {\bf72.6} &{\bf 86.5} \\
\hline
\end{tabular}
}
\vspace{-0.45cm}
\label{tab:tab1}
\end{table}

\begin{table}[t]
\centering
\caption{Recognition accuracy of radar emitters with different PRI modulation on $P_4$.}
\setlength{\tabcolsep}{1.9mm}{
\begin{tabular}{lllllll}
\hline
Methods                     & CST       & JIT       & SLD       & WOB       & D\&S      &STG        \\
\hline
CNN \cite{2018LiCNN}        & 35.7      & 30.4      & 57.4      & 60.8      &51.5       &49.4       \\
A-RNN \cite{2020LiRNN}      & 40.4      & 40.6      & 59.3      & 57.7      &53.0       &50.8       \\
ACSE \cite{2020ACSE}        & 40.0      & 38.6      & 53.4      & 58.6      &54.5       &49.0       \\
TCN \cite{2021TCN}          & 46.4      & 48.3      & 56.3      & 54.9      &58.5       &51.6       \\
ERM \cite{ERM}              & 49.8      & 54.6      & 62.1      & 64.0      &59.7       &56.8       \\
d-SNE \cite{2019d-SNE}      & 56.6      & 55.9      & 65.3      & 70.7      &62.8       &60.8       \\
MADA \cite{2020MADA}        & 58.5      & 56.1      & 64.6      & 62.0      &59.2       &59.2       \\
Ours                        & {\bf67.9} & {\bf59.1} & {\bf78.3} & {\bf85.0} &{\bf81.0}  &{\bf70.9}  \\
\hline
\end{tabular}
}
\vspace{-0.45cm}
\label{tab:tab2}
\end{table}

\begin{table}[t]
\centering
\vspace{-0.25cm}
\caption{Recognition accuracy of radar emitters with STG PRI but different PRI values on $P_4$.}
\begin{tabular}{llllll}
\hline
Methods                     & STG1      & STG2      & STG3      & STG4      & STG5      \\
\hline
CNN \cite{2018LiCNN}        & 51.0      & 56.8      & 47.7      & 43.5      &49.6       \\
A-RNN \cite{2020LiRNN}      & 52.8      & 51.1      & 45.1      & 49.1      &56.1       \\
ACSE \cite{2020ACSE}        & 50.4      & 53.2      & 49.5      & 47.9      &45.1       \\
TCN \cite{2021TCN}          & 56.3      & 53.6      & 47.0      & 47.0      &54.8       \\
ERM \cite{ERM}              & 62.5      & 51.9      & 51.4      & 56.0      &63.3       \\
d-SNE \cite{2019d-SNE}      & 62.8      & 60.9      & {\bf61.7} & 57.2      &63.4       \\
MADA \cite{2020MADA}        & 63.5      & 61.7      & 53.2      & 56.2      &62.8       \\
Ours                        & {\bf76.1} & {\bf74.7} & {\bf61.7} & {\bf69.1} &{\bf73.0}  \\
\hline
\end{tabular}
\vspace{-0.45cm}
\label{tab:tab3}
\end{table}

Table \ref{tab:tab1} compares our method against state-of-the-art baseline on the standard dataset. As reported in the table, on $P_1$, $P_3$, and $P_4$, our method outperforms the competitor methods. Specifically, our method outperforms A-RNN by 10.1\%,  ACSE by 10.6\%, and TCN by 9.5\%. It also outperforms d-SNE by 6.4\% and MADA by 4.9\%. On $P_2$, our method's performance is comparable to the competitor schemes because $P_2$ has the same error ratio as the training set. The results demonstrate that as the missing and spurious pulses ratio increases, the other models can hardly recognize the radar emitter accurately because the source domain is substantially different from the target domain. When the EW environment is the worst, our method affords a 20.4\% performance improvement over TCN and 11.2\% over MADA. This is because our method generates reliable emitter signals and forces the model to learn the cross-domain features, enhancing its adaptability to emitter signals in changing environments.

Table \ref{tab:tab2} compares the recognition accuracy of our method against the baselines on different PRI modulations on $P_4$. Our method achieves better accuracy than the other methods on all PRI types. However, our method's deficiency is that the recognition accuracy of constant PRI and jittered PRI is still unsatisfactory because these are easily confused under high error ratios. Table \ref{tab:tab3} compares the recognition accuracy of five staggered PRI on $P_4$, where the suggested method achieves the best performance among all baselines on each emitter. Both tables prove that our method can well recognize the emitters with different or the same PRI modulations.

\vspace{-0.3cm}
\subsection{Evaluation of Few-shot Domain Adaptation}
\vspace{-0.3cm}

\begin{figure}[ht]
\begin{minipage}[b]{1.0\linewidth}
  \centering
  \centerline{\includegraphics[width=9cm]{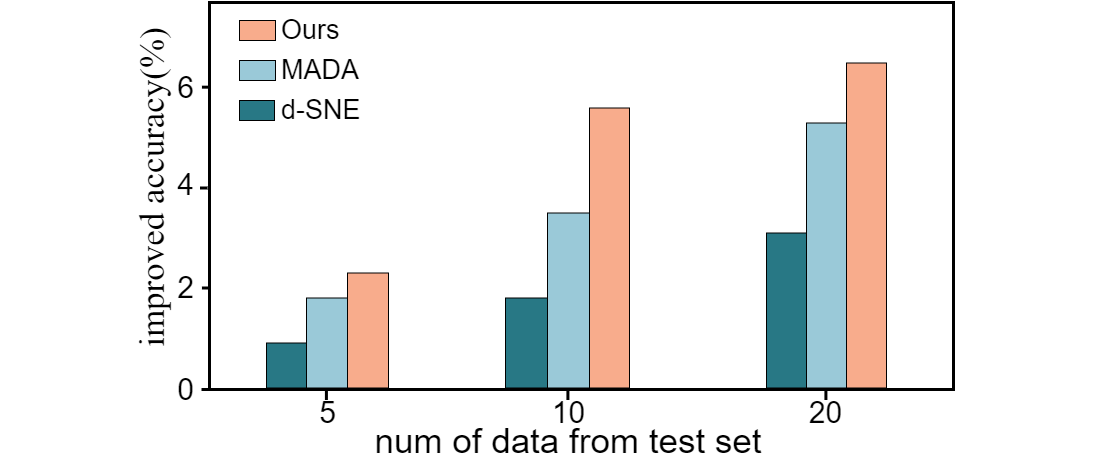}}
\end{minipage}
\vspace{-0.8cm}
\caption{Results of using a few samples from the testing set for training.}
\label{fig:results_oth}
\vspace{-0.3cm}
\end{figure}

Finally, we add a few samples from unlabeled target data for training.  For this trial, we use the existing DG methods with Fig.\ref{fig:results_oth} illustrating that fine-tuning with a few samples from the target domain can significantly improve the model's performance on the target domain. Our model performs better than d-SNE and MADA, but as the data from the testing set increase, the improved accuracy advantage gradually decreases. Since the signals generated by our generator present enough diversity, it is difficult to improve the recognition accuracy with superfluous samples from the testing set.

\vspace{-0.2cm}
\section{Conclusion}
\label{sec:Conclusion}
\vspace{-0.2cm}

This paper proposed a novel domain generalization framework to improve the adaptability of signal recognition in changing environments. 
Our method first designed several noise generators which simulate varied scenes accordingly to enhance the diversity of the detected signals but keep the semantic features of the signals themselves. In addition, the suggested model can learn the cross-domain signal feature through adversarial learning. Extensive experiments on the standard dataset demonstrate that our method achieves the best radar emitter signal recognition performance in changing environments. Our method fills the gap in the task and provides a promising direction to solve the radar emitter signal recognition problem.

\clearpage
\vfill\pagebreak

\bibliographystyle{IEEEbib}
\bibliography{strings,Template}

\end{document}